\documentclass[journal]{IEEEtran}
\usepackage{cite}
\usepackage{multirow}
\usepackage{graphicx}
\usepackage{supertabular}
\usepackage{longtable,booktabs}

\ifCLASSINFOpdf
\else
\fi
\usepackage{amsmath}
\begin{document}
%
\title{Automatic Image Co-Segmentation: A Survey}
%
%
%

\author{Xiabi Liu*,Xin Duan 
\thanks{Xiabi Liu* (corresponding author) is with Beijing Lab of Intelligent Information
	Technology, School of Computer Science, Beijing Institute of Technology,
	Beijing 100081, China (e-mail: liuxiabi@bit.edu.cn).}
\thanks{Xin Duan was admitted to the school of Computer at Beijing Institute of Technology, Beijing, China. She is currently working in the MLMR Lab at Beijing Institute of Technology. }}

\maketitle

\begin{abstract}
Image co-segmentation is important for its advantage of alleviating the ill-pose nature of image segmentation through exploring the correlation between related images. Many automatic image co-segmentation algorithms have been developed in the last decade, which are investigated comprehensively in this paper. We firstly analyze visual/semantic cues for guiding image co-segmentation, including object cues and correlation cues. Then we describe the traditional methods in three categories of object elements based, object regions/contours based, common object model based. In the next part, deep learning based methods are reviewed. Furthermore, widely used test datasets and evaluation criteria are introduced and the reported performances of the surveyed algorithms are compared with each other. Finally, we discuss the current challenges and possible future directions and conclude the paper. Hopefully, this comprehensive investigation will be helpful for the development of image co-segmentation technique.
\end{abstract}


%
\IEEEpeerreviewmaketitle

\section{Introduction}
%
%
%
%
\par Image co-segmentation is a problem of segmenting common and salient objects from a set of related images. Since this concept was firstly introduced in 2006\cite{rother2006cosegmentation}, it has attracted a lot of attentions and many co-segmentation algorithms have been proposed. The reasons behind its importance are two folds. On the technique aspect, the correlation between images brings valuable cues for defining the interested objects and alleviates the ill-pose nature of segmentation. On the application aspect, image co-segmentation algorithms can be applied to and play crucial roles in various applications, such as Internet image mining, image retrieval, video tracking, video segmentation, and etc. 
\par The visual/semantic object features in images plus the correlation between images provide needful cues for achieving the goal of co-segmentation. Most of previous co-segmentation algorithms explored handcrafted object features and correlations and conducted the computation on the level of object elements (such as pixels, superpixels, or over-segmented regions), object regions/contours, or common object models.

\begin{itemize}
	\item For object elements, we usually establish an objective function, mostly often in form of energy function, to describe the intra-image and the inter-image relationship among object elements. By optimizing such objective function, we determine the class labeling of each element in each image.
	\item Object regions/contours are corresponding with the whole objects in each image, or in other words, the segmentation results in each image. A model is used to represent object regions or contours. We try to find out the optimal models fitted to the images for completing the segmentation. In such processing, object regions can be explored alone, while object contours are usually combined with regions.
	\item Common object models are used to model the common objects across the images, which keep same for the entire correlated images while object regions/contour models are specific for single image and vary across the images. We try to find out an optimal common object model according to all the explored images. 	
\end{itemize}

\par There is a hierarchical relationship among the three factors above. Actually, an object region/contour is composed of object elements, while a common object model generates object regions/contours in each image. 
\par Although much progress has been made, the algorithms based on handcrafted object features and correlations still suffer from their unrobustness and inexactitude. Recently, deep learning is introduced to improve the performance of image co-segmentation through mining more sophisticated object features and correlations from data. In the following, we call deep learning based image co-segmentation methods as deep methods for short, and other methods as non-deep methods.
\par In this paper we survey automatic image co-segmentation algorithms, including non-deep and deep ones. We only consider fully automatic algorithms. Those involving user interactions are ignored. Firstly, we describe visual/semantic object cues and correlation cues for guiding the image co-segmentation. Secondly, we introduce main computation frameworks behind non-deep methods, followed by the descriptions of non-deep methods based on object elements, object regions/contours, and common object models, respectively. Thirdly, the deep methods are reviewed. We then discuss the evaluation of algorithms and summarize the performance of surveyed algorithms. Finally, the current challenges and possible future directions are discussed, and the conclusions are made. We think that such comprehensive investigation of image co-segmentation is helpful to the development of this technique.

\section{Segmentation Cues}

\par No matter whether non-deep methods or deep ones are used, visual/semantic object cues and correlation cues for discriminating interested objects from backgrounds are the basis of getting desired co-segmentation results. We mean the characteristics of objects observed on each single image by “object cues”, which include the difference between foreground and background, the smoothness of foreground/background, the saliency and the objectness of objects. Contrarily, the correlation cues are corresponding with multiple images, mainly the commonness and the co-saliency of common objects, which brings the main advantage of alleviating the ill-pose nature of co-segmentation, compared with other segmentation problems.

\subsection{Similarity}
\par For a good segmentation, the elements in each segmented object or the objects as a whole should be similar with each other, while those for different objects should be different from each other. Here the concept of “object” include the background.
\par A low-level visual similarity is measured in visual features of elements. The two commonly used visual similarity measurements are L2/L1 distance and cosine similarity. The former is used more often in non-deep methods and is usually transformed to a similarity value based on Gaussian model\cite{dai2013cosegmentation,joulin2010discriminative,kim2011distributed,lee2015multiple,rother2006cosegmentation,yuan2017deep,han2018robust,liu2014object,chang2011co,liu2017image}. The cosine similarity can be used in non-deep methods\cite{li2018deep,tao2017image} and can play more important role in deep methods\cite{li2018deep}, since it is much easier to be implemented in convolutional form. For computing the similarity between object components across the images, the matching of object components can be employed\cite{faktor2013co,taniai2016joint,zhang2014image,rubinstein2013unsupervised,kim2011distributed,meng2016cosegmentation} or not.
\par The higher-level similarity is measured in semantics. For complicated objects with various appearances, we need to consider their semantics and observe the possibilities of having corresponding semantics for objects or the elements in objects. The semantic modeling tools for completing such task include Gaussian Mixture Models (GMMs)\cite{dai2013cosegmentation,meng2016cosegmentation,jerripothula2017object}, histogram\cite{meng2013image,rother2006cosegmentation,rubinstein2013unsupervised,hochbaum2009efficient,chang2011co,mukherjee2009half,kim2012hierarchical,mukherjee2011scale}, Dense Conditional Random Fields (DCRF)\cite{yuan2017deep}, classifiers such as support vector machines (SVM)\cite{rubio2012unsupervised,liang2017multi,Jian2013Learning,chai2012tricos}, random forest classifier\cite{zhu2014multiple}, and neural networks\cite{han2018robust}, and functions such as active basis model\cite{dai2013cosegmentation}, logistic function\cite{li2016object}, deformable part model\cite{liu2014object}, random forest regressor\cite{vicente2011object}, geometric transformation\cite{taniai2016joint}, functional map\cite{wang2013image}, and spatial pyramid matching  (SPM)\cite{kim2012multiple}.  We can use these models to represent the objects in a single image, as well as the common objects across the images.

\subsection{Smoothness}
\par It is reasonable and expected to obtain visually smooth results of segmentation, which is usually achieved by constraining that neighboring and similar elements should have the same semantic labels. In order to realize this goal, we often try to maximize the similarities between the features of neighboring and same labeled elements\cite{dai2013cosegmentation,lee2015multiple,han2018robust,jerripothula2017object,rubinstein2013unsupervised,hochbaum2009efficient,liu2017image,taniai2016joint,Jian2013Learning,faktor2013co,kim2012hierarchical,wang2013semi}, or minimize the similarities between the features of neighboring but differently labeled elements\cite{rother2006cosegmentation,chang2011co,mukherjee2009half}.

\subsection{Saliency}	
\par In many cases, visually salient parts in the images correspond to the objects we are interested in. The concept of saliency is inspired by the human visual attention mechanisms, thus is also often called visual attention modeling, which can be defined in two ways: based on human eye fixation prediction and based on visual saliency in the context. According to Borji and Itti\cite{borji2013state}, the saliency computation approaches can be divided into bottom-up methods, top-down methods, and neural network based methods. The top-down models can be further divided into visual search methods, context methods, and task-driven methods. 
\par Co-saliency tries to identify informative and salient regions in a group of related images. The results of co-saliency are obviously helpful for co-segmentation. According to Zhang et al.\cite{zhang2018review}, the co-saliency methods can be divided into three categories, bottom-up, fusion-based, and learning-based.

\subsection{Objectness}
\par Objectness denots the objects’ characteristics that make them discriminated from non-objects such as sky, sea, walls, and such things. These non-objects are called ‘stuff’. So the objectness and the stuff actually construct a specific binary semantics. We can measure the objectness scores of the elements in an image using the binary classifiers\cite{vicente2011object,fu2015object,meng2012object}.

\section{Graph based Optimization Frameworks}

\par Many non-deep co-segmentation algorithms were developed under graph based optimization frameworks. Four commonly used frameworks include graph cuts\cite{boykov1999fast}, GrabCut\cite{rother2012interactive}, normalized cuts\cite{shi2000normalized}, and random walks\cite{grady2006random}. Under these frameworks, each image is transformed to a graph. Nodes in the graph are divided into two types. The first type is corresponding with the elements in the segmentation results. The second type is corresponding with the segmentation labels, i.e., object categories. These two types of nodes are called element nodes and class nodes in the following, respectively. Edges in the graph reflect the relationship between nodes. For two element nodes, we usually consider the visual similarity or smoothness between them. For an element node and a class node, we usually consider corresponding semantic similarity.

\subsection{Graph Cuts}
\par In graph cuts, we compute the minimum cut on the graph, which leads to the minimization of an energy function representing the segmentation objective and divide the graph into unconnected subgraphs. Each subgraph includes one and only one class node and thus form a segmented result for specific object class. The minimum cut can be computed very efficiently by max flow algorithms. In order to apply graph cuts technique, the energy function should be defined as submodular or called regular. Kolmogorov and Zabih\cite{kolmogorov2004energy} defined the regular energy functions that can be minimized by the graph cuts and presented a method of graph construction for any regular functions.
\subsection{GrabCut}
\par The GrabCut is a widely used extension of the graph cuts. The main characteristic of it is to introduce GMM to model the distribution of elements’ features for each segmentation label. The GMM for each label is trained from the examples belonging to this category. Then the GMMs are used to compute the classification confidences for each category and use them as the weights of the edges between element nodes and corresponding class nodes. On such constructed graph, the graph cuts is performed to get the segmentation results. And the segmentation results are returned to retrain the GMMs. These two procedures, GMM training and graph cuts based segmentation, are alternated until the computation is stable or the maximum iteration is attained.

\subsection{Normalized Cuts}
\par In normalized cuts, the graph is represented as a matrix of edge weights. A Laplacian matrix is further constructed by subtracting the weight matrix from a diagonal mass matrix. The values in the diagonal of mass matrix are the total edge weights connecting to the corresponding nodes. Let {\bfseries W} be the weight matrix, {\bfseries D} be the diagonal mass matrix. We solve an eigensystem: 	
\begin{equation}
D^{\frac{1}{2}}\left (D-W\right )D^{\frac{1}{2}}x= \lambda x
\end{equation}
where {\bfseries x} is the vector indicating the binary segmentation of each node. The second smallest eigenvector of such eigensystem gives the segmentation result that minimizes cut costs normalized by the total weight of edges. The solution can be extended to multiple object segmentation by recursively performing this binary segmentation on two partitioned parts.
\par Normalized cuts are developed based on spectral graph theory and belong to spectral clustering technique, under which we can develop other closely related segmentation methods such as average cost\cite{shi2000normalized}, discriminative clustering\cite{joulin2010discriminative}, and etc.

\subsection{Random Walkers}
\par Under random walkers framework, some element nodes are labeled with corresponding segmentation labels. Then for each unlabeled element node, we can imagine that a random walker start walking from it and the probabilities that the walker first reaches any one of labeled nodes can be computed. The labeling of the node with the maximum probability is taken as the result of this unlabeled element node. Such probability computation problem can be transformed to the combinatorial Dirichlet problem and solved by solving a corresponding system of linear functions under the constraint that the probabilities at any node will sum to unity\cite{grady2006random}. Single random walker can be further extended to multiple random walkers that walk simultaneously and interact with each other to get desired results\cite{lee2015multiple}.

\section{Object elements based non-deep co-segmentation}
\par Object elements include pixels, superpixels, and over-segmented regions. Superpixels and over-segmented regions are used more often than pixels for practical computation. By considering the segmentation cues described in Section 2 under the optimization frameworks described in Section 3, we obtain the corresponding object elements based algorithms.

\subsection{By GrabCut}
\par Li et al.\cite{li2018unsupervised} computed the saliency of each pixel in each image and used them to rank the images. According to the ranking and the similarities between hierarchical over-segmented regions across the images, they selected the simple images that contain common objects. These simple images are co-segmented by exploring the similarities between over-segmented regions across the images under GrabCut optimization framework. Finally, the co-segmented results of foreground and background in simple images are used as positive and negative examples to guide the segmentation of remaining complex images.
\par Faktor and Irani\cite{faktor2013co} thought that the co-segmentation results should share large non-trivial regions. Thus they firstly detected co-occurring regions across the images by using the randomized search and the propagation algorithm. Then the hierarchical segmentation was performed on each image separately to get the initial segments. According to the overlapping degree of initial segments and co-occurring regions, the co-segment scores were computed for each pixel. After smoothing such co-segment score map over the neighboring pixels in a same image and over the corresponding pixels in co-occurring regions, the GrabCut was performed on the co-segment score map to get the final segmentation result.
\par Liang et al.\cite{liang2017multi} got co-saliency values of superpixels in each image. Then they trained a SVM according to the co-saliency ranking for all the images. This SVM was used to update the co-saliency map. Finally, the GrabCut was performed on the co-saliency map to get the final result.

\subsection{By Normalized Cuts}
\par Ma et al.\cite{ma2017unsupervised} firstly obtained structure-meaningful regions based on the similarities between superpixels. Then an L1-manifold graph was constructed for all the structure-meaningful regions in all the images to express implicit coherency across the images. And a graph for describing relationship among all the regions in each image was established separately, according to the similarities between regions. The two graphs were combined to form a hyper-graph. The normalized cuts was performed on this hyper-graph to get initial segmentation on the level of superpixels. Then the GrabCut was used to refine the results on the level of pixels.
\par Kim et al.\cite{kim2012hierarchical} presented a multi-scale normalized cuts method. Firstly, they clustered the images in the database to groups and do co-segmentation on each group separately. For the co-segmentation on each group, the gpb-owt-ucm method was employed to obtain hierarchical segmentation in each image, respectively. Then the normalized cuts was conducted to complete the computation, which is based on the affinity between neighboring superpixels in each hierarchy, the constraint for parent-child superpixels in adjacent layers of the hierarchy, and the affinity between superpixels from two images.

\subsection{Random Walker}
\par Kim and Xing\cite{kim2011distributed} firstly computed the clustering centers in each image by using anisotropic diffusion method, and then completed the segmentation by using random walker. A similar approach was reported in\cite{liu2017image}.
\par Lee et al.\cite{lee2015multiple} firstly obtained a graph for each image according to the correlation among superpixels. Then the multiple random walker algorithm was performed on this graph to obtain foreground and background distribution. Then the foreground and background distributions were refined by the bilateral filter based on the relationship between images. Finally, the segmentation was decided according to maximum a posterior criterion.

\subsection{Clustering Based}
\par Tao and Fu\cite{tao2017image} obtained the partition-level side information and visual features of superpixels. Then SGC3 clustering is performed on such descriptions of superpixels in all the images to get two clusters as the segmentation result.
\par Joulin et al.\cite{joulin2010discriminative} established a computation matrix according to the space consistency of pixels in each single image and the feature similarities between pixels from different images. Then the spectrum clustering was performed on the computation matrix by using efficient convex relaxation algorithm to obtain the segmentation result.

\section{Object regions/contours based non-deep co-segmentation}
\par In such category, the regions and/or contours for the whole objects in each image are represented and the relationships between regions/contours across the images, usually the similarities between them, are explored. The regions can be used alone and can be represented by the histogram, the GMM, classifiers, or features. The contours are usually combined with regions for completing the computation. 

\subsection{Histogram based}
\par Wang and Liu\cite{wang2013semi} used color histograms to represent foreground regions. The distance between histograms from different images, the label smoothness for superpixels in each image, and the constraint on the ratio of foreground superpixels to background ones are combined to establish a binary quadratic programming problem that is optimized by using the trust region based algorithm. 
\par Hochbaum and Singh\cite{hochbaum2009efficient} considered the fitting degrees of foreground pixels to the foreground model, the label smoothness, and the similarity between foreground histograms of two images to define an energy function, which is minimized by the graph cuts to get the segmentation. 
\par Chang et al.\cite{chang2011co} defined an energy function based on the pixel saliency, the label smoothness, the difference between foreground and background histograms in each image, the difference between the foreground histograms from two images, and the co-occurrence of pixels on different images. The graph cuts was employed for minimizing this energy function to complete the segmentation. 
\par In the method of Muhkerjee et al.\cite{mukherjee2009half}, the classification cost of pixels, the label smoothness, and the difference between two foreground histograms of two images were combined to form an energy function and solved by the graph cuts. 
\par The energy function of Roth et al. \cite{rother2006cosegmentation} was composed of the label smoothness, the number of foreground pixels, and the similarity between foreground histograms of two images. The minimization of this energy function was initialized by using the submodular-supermodular procedure and refined by using the trust region graph cuts.
\par Rubinstein\cite{rubinstein2013unsupervised} defined an energy function by employing the saliency and the sparsity of foreground pixels and the consistency of foreground pixels between images. To solve the energy function’s minimization, the segmentation of each image was optimized alternately.

\par Taniai et al.\cite{taniai2016joint} explored the fitting of color histograms of foreground and background to superpixels, the label smoothness, the similarities between matched pixels in two images, and the smoothness of geometric transformation from the matching result to define an energy function. Furthermore, a hierarchical graph was constructed to represent the multi-scale computation, on which the bottom-up procedure and top-down procedure were conducted to solve the problem.

\subsection{GMM based}
\par Meng et al.\cite{meng2016cosegmentation} divided the images into several groups according to the correlation among them. Then the energy was defined by considering the single-image consistency based on GMM modeling, the single-group consistency based on GMM modeling, and the multi-group consistency based on SIFT flow. The minimization of the energy is realized by using the Expectation-Minimization algorithm. In the expectation step, the correspondence between segmentation regions from different images was updated according to the current segmentation results. In the minimization step, the region segmentation results were updated by using the graph cut according to the correspondence between segmented regions. 
\subsection{Classifier based}
\par Cech et al.\cite{cech2010efficient} alternately optimized the correspondence of two images and the co-segmentation in them. According to the correspondence result, the seed points were determined and grow up to form the co-segmentation regions as well as the new correspondence result. Then this procedure was repeated on the new correspondence result. A support vector machine (SVM) was introduced and trained on the training dataset to evaluate the correspondence result.

\subsection{Features Based}
\par Fu et al.\cite{fu2015object} introduced the pixel depth information for improving co-segmentation accuracy. They selected the object regions in all the images from the initial candidates obtained by using 2.5D region proposals generation method. The optimal selection was expressed as an integer quadratic programming problem by considering the objectness and the co-saliency of candidates, the distance between two candidates, and the mutex constraints for overlapping regions. Such integer quadratic programming problem was solved by using the fixed-point iteration algorithm.
\par Meng et al.\cite{meng2012object} used three different methods (superpixel, saliency, object detection) to get the local regions. Then a graph was constructed according to the similarities between local regions, the saliency of local regions, and the sequence of the images. By finding the shortest path on such graph, the segmentation result was obtained.

\subsection{Combining Regions and Contours}
\par Dai et al.\cite{dai2013cosegmentation} expressed the object contours by the sketch models and described the object regions by the color distributions of pixels in it. Then they defined an energy function composed of three terms: for the sketch models, for the region models, and for the consistency between the object contours and the object regions. They firstly minimized the energy of each single image by alternately optimizing the object contour and the object region, and then minimized the sum of the energies of all the images by sequentially optimizing the object contour models, region color distributions, and segmentation templates. 
\par Li et al.\cite{jerripothula2017object} alternately optimized the region segmentation result and the skeletonization result by utilizing the correlation between them. The computation of skeleton under given segmentation result and the optimization of segmentation result based on the skeleton were both completed through energy function minimization, respectively. 
\par Meng et al.\cite{meng2013image} applied the level set to represent and optimize the object contours in each image. The corresponding energy function was established by considering the length of contours, the size of object regions in contours, the fitting degrees of background pixels to the background histogram in each image, and the fitting degrees of foreground pixels to the foreground histogram in another image.

\section{Common object model based non-deep co-segmentation}
\par In this category, a common object model should be constructed to describe those common objects across the images. The mainly used object models include the histogram, GMM, classifiers, and functions.
\subsection{Histogram based}
\par Mukherjee et al.\cite{mukherjee2011scale} represented foreground regions using histograms and tried to obtain similar foreground histograms across the images. They transformed this computation goal to be that the rank of the matrix constructed from all the images’ histograms should be 1. Such transformed objective was solved by introducing slack constraints and alternating optimization between matrix update and segmentation update.
\subsection{GMM based}
\par Kim and Xing\cite{kim2012multiple} considered multi-category common objects’ co-segmentation. They applied GMM and spatial pyramid matching (SPM) with linear SVM as the model of each object category. The common object models and the classification of over-segmented regions to corresponding object models were alternately optimized until the classification confidences of over-segmented regions were maximized.

\subsection{Classifier based}
\par Chai et al.\cite{chai2012tricos} employed two types of SVM classifiers, category-level SVM and dataset-level SVM. The category-level SVM was used to classify foreground /background, which was trained on the labeling results of superpixels from the GrabCut over the saliency measurement. The SVM training and superpixel labeling were updated alternately. The dataset-level SVM was used to classify foreground regions into the object categories, which was trained on the segmented regions in previous step and returned to refine labeling results of superpixels.
\par Sun and Ponce\cite{Jian2013Learning} trained a part detector based on SVM with group sparsity regularization for detecting the common object appearing in the images, from a lot of initial detectors. Based on such SVM, an energy function was established by considering the discriminative clustering in each image, object cues for all the images, and the labeling smoothness. The energy is optimized by using the graph cuts. An alternating optimization between SVM training and graph cuts based segmentation labeling was conducted to solve the co-segmentation.
\par Zhu et al.\cite{zhu2014multiple} considered four information to define an energy function: pixel classification confidence based on random forest classifier, superpixel consistency based on random forest classifier, object region consistency, and the correlation between two pixels. This energy function was minimized by using the graph cuts. 
\par Rubio et al.\cite{rubio2012unsupervised} considered the following four components to establish an energy function: the probability of pixel being foreground or background, the probability of over-segmented regions being foreground or background, the consistency between pixels and over-segmented regions in each image, and the consistency for over-segmented regions between images. SVM and GMM were used to model foreground and background regions. And the graph cuts was employed to minimize the energy. 
\par Zhang et al.\cite{zhang2014image} obtained the initial segmentation based on the saliency computation. Then the multi-task learning was employed to obtain the linear classifier for discriminating the foreground from the background. Finally, the classifier was used to select the seed superpixes, based on which the GrabCut was used to complete the segmentation. 
\par Vicente et al.\cite{vicente2011object} obtained the initial segmentation results for each image and then constructed an objective function based on the similarities between initial segmentation results on images. Here the similarity was computed by using random forest regressor. Finally, they used A* search method to optimize the objective function to get the final results.

\subsection{Functions Based}
\par Wang et al.\cite{wang2013image} described the segmentation result as a binary indicator function, which is represented as the weighted sum of basis functions. Then the relationship between the segmentation of correlated images (defined as similar in features) was computed as the map of corresponding functions. Such map for all the correlated images should satisfy the constraint of cycle consistency. Under this constraint, the best function map was obtained and used to measure the map consistency term for correlated images. The authors tried to find the segmentation leading to the minimum sum of this map consistency term and the segmentation term from using normalized cuts on each single image. Such objective was solved by using normalized cuts again to get the final result. 
\par Li et al.\cite{li2016object} found out salient and common regions as the segmentation result. Such objective was described as the sum of three sub-objectives: 1) segmentation results should be salient; 2) segmentation results should fit to the common object; and 3) the saliency detection result should be consistent with the common object detection result. Accordingly, the following three procedures were conducted alternately to achieve the overall objective: 1) detecting the saliency in each image by using low rank matrix recovery; 2) learning the logistic function for superpixels belonging to the common foreground; and 3) updating the probabilities of superpixels being common foreground based on saliency detection result and learned logistic function.

\subsection{Others}
\par Liu et al.\cite{liu2014object} computed the energy of each pixel based on deformable part model learned from the training data and the energy for reflecting the labeling smoothness. The sum of these two energies were minimized by using the graph cuts to get the segmentation result.

\section{Deep co-segmentation}
\par Recently, we witness the applications of deep networks to the co-segmentation. Although the number of works in this category is still much less than traditional non-deep ones and most of such works embedded deep networks into traditional methods, they showed that this is a promising direction to be explored further. 
\par Yuan et al.\cite{yuan2017deep} used a deep network to describe the dense conditional random fields (DCRF) for the common object in the images. Such deep network was used to compute the probability of each pixel being the foreground, which was used in the second DCRF procedure to decide the final labels of pixels. 
\par Wang et al.\cite{wang2017multiple} used fully convolutional network (FCN) to obtain the initial segmentation result on each single image. Then the con-occurrence of candidate regions was computed by using N-partite graph method. Finally the GrabCut was applied on the con-occurrence map to obtain the segmentation results.
\par Quan et al.\cite{han2018robust} constructed two graphs according to low-level visual features and high semantic features. The high semantic features were computed by using convolutional neural network (CNN). On these two graphs, two probability maps corresponding to the segmentation results were computed. Finally the GrabCut was performed on the multiplication of these two probability maps to obtain the final result. 
\par Li et al.\cite{li2018deep} applied Siamese network to co-segmentation, which is composed of three parts: Siamese encoder for using CNN network to extract features from the two images, the mutual correlation for matching two image features, and Siamese decoder for obtaining segmentation results by using the deconvolutional computation. To our knowledge, this work is only one totally dependent on deep networks and can be trained end-to-end.

\section{Algorithm Evaluations}
\par The performance evaluation of the algorithms is important for the development of the technique, which can provide an understanding of the state-of-the-arts and help to analyze the strengths and the weaknesses of each algorithm. For this purpose we need evaluation datasets and criteria.
\subsection{Datasets}
\par Currently, the widely used datasets for evaluating co-segmentation algorithms include iCoseg\cite{batra2010icoseg}, MSRC\cite{shotton2006textonboost}, FilckerMFC\cite{kim2012multiple}, and Internet\cite{rubinstein2013unsupervised}.

\begin{itemize}
	\item {\textbf iCoseg dataset\footnote{http://chenlab.ece.cornell.edu/projects/touch-coseg/}} contains 643 images divided into 38 object groups. Each group contains 17 images on average. The pixel-wise hand-annotated ground truth is offered. The backgrounds in each group are consistent natural scenes. Instead of entire iCoseg dataset, many previous works used various versions of its subsets. For example, a subset of 122 images with 16 classes was often used. In the following, we indicate the class number in the title of iCoseg datasets, such as iCoseg-38, iCoseg-16, and etc.
	
	\item  {\bfseries MSRC\footnote{http://www.research.microsoft.com/en-us/projects/objectclassrecognition/}} is composed of 591 images of 23 object groups. The ground truth is roughly labeled, which does not align exactly with the object boundaries. The backgrounds in each group are natural scenes with a few clutters. Like the situation in iCoseg, the subsets of MSRC were often used, especially a subset with 14 classes. We also indicate the class number in the title of MSRC datasets in the following descriptions, such as MSRC-14, MSRC-7, and etc. Furthermore, Weizman Horses dataset\footnote{http://www.msri.org/people/members/eranb/}  was often mix-used with MSRC-7. We call such dataset as MSRC-7+1. 
	
	\item {\bfseries FilckerMFC} is designed for multiple foreground co-segmentation, i.e., jointly segmenting more than one foreground objects from input images. This dataset consists of 14 groups with 10～20 images/group. Each group contains a ﬁnite number of foreground objects, and only a subset of foreground objects are contained in each image. The pixel-level ground truth is provided.
	
	\item {\bfseries Internet\footnote{http://people.csail.mit.edu/mrub/ObjectDiscovery/}} consists of 3 classes (airplane, car, and house) of thousands of images downloaded from Internet. A subset of it with randomly selected 100 images was more often used. We call it as Internet-100, while the entire dataset as Internet-entire.
	
\end{itemize}

\par Except the four datasets above, ImageNet\cite{deng2009imagenet}\footnote{http://www.image-net.org/}, PASCAL VOC 2010 and 2012\cite{everingham2015pascal}\footnote{http://host.robots.ox.ac.uk/pascal/VOC/} (simplified as PASCAL VOC in the following), and Oxford flowers dataset\cite{nilsback2006visual}\footnote{http://www.robots.ox.ac.uk/~vgg/data/flowers/} are also considered in the experiments of co-segmentation. ImageNet contains a huge number of examples. However, only boxes bounding the objects are provided, thus it is not very exact to test the effectiveness of segmentation algorithms on ImageNet. The people often use it to inspect the scalability of the algorithms, considering its size. PSACAL VOC datasets were frequently appeared in the works of deep co-segmentation. The number of annotated examples in iCoseg, MSRC, FlickerMFC, or Internet is limited and is not enough for training machine learning based methods, especially for deep learning based ones. PSACAL VOC datasets with more than 10k annotated images can be used to alleviate this problem. Thus it is usually used as the training set, instead of the test set. As for Oxford flowers dataset, it involves only one object category and was not used very often in image co-segmentation community.

\subsection{Evaluation Criteria}
\par The evaluation criteria used in previous co-segmentation works include pixel accuracy (PA), intersection over union (IOU), F score, globe region covering error, variation of information, and probabilistic random index. 
\par Among these criteria, PA, IOU and F score measure the segmentation quality from the view of pixel, region, and contour, respectively. We think that such three perspectives of evaluation are enough. However, the evaluations of previous algorithms are mainly dependent on PA and IOU; F score was seldom used. So we described these three criteria in the following, but collect the performance of previous algorithms on only PA and IOU. 
\begin{itemize}
	\item{\bfseries Pixel Accuracy (PA)} measures the segmentation precision by computing the rate of mislabeled pixels. Let {\bfseries $W$} be the number of wrongly labeled pixels, {\bfseries $T$} be the total pixels in an image, then the pixel accuracy is $W/T$.
	\item{\bfseries Intersection Over Union (IOU)} is also well known as Jaccard index. Different form PA, it is region based, instead of inspecting each pixel separately. Let $M$ and $G$ be the sets of foreground pixels in the segmented image and in the ground truth, respectively, then the IOU value is measured as $\frac{M\cap G}{M\cup G}$.
	\item{\bfseries F score} captures the trade-off between the precision rate and the recall rate, measured on the segmentation labeling of boundary pixels. If a boundary pixel in the ground-truth is detected in the segmented image, we call it “true positive”; otherwise, we call it “false negative”. The meanings of “true negative” and “false positive” are defined similarly. Let TP, TN, FP, FN be the number of true positives, true negatives, false positives and false negatives, respectively. Then the precision rate (P) and the recall rate (R) are measure as $P=\frac{TP}{TP+FN}$ and $R=\frac{TP}{TP+FP}$,respectively. F score is the harmonic mean of precision and recall, which can be defined as $\frac{2PR}{P+R}$.
	
\end{itemize}

\subsection{Performance Comparisons}
\par We collect the values of PA and IOU for image co-segmentation, reported in previous works on evaluation datasets, and list them in Table 1. All the shown results are the average ones on all the classes in the datasets, provided by the authors or calculated over the results for each class provided in the papers. The values are accurate to the third decimal place. PA is shown in the form of percentage, while IOU in the form of decimal. Furthermore, some works display the results in figures, but don’t provide the exact values. We determine the upper bounds of values shown in the figures and report them in Table 1. 
\par In order to demonstrate the current state-of-the-art performance more clearly, we display the best PA and IOU values over various test sets in Fig.1. According to Table 1 and Fig.1, we can conclude that

\begin{itemize}
	\item The performances of current co-segmentation algorithms are still not very satisfactory. The best PAs are 96\%, 94.4\%, <60\%, and 93.3\% for iCoseg, MSRC, FlickerMFC, and Internet, respectively. And the best IOUs are even less to 0.860, 0.799, 0.647, and 0.704 for these four datasets, respectively.
	\item There are so many different versions of datasets used in the experiments. We have encountered 10 iCoseg, 5 MSRC, and 2 Internet. Even so, we still cannot get really good segmentation effects. This phenomenon further proves the difficulty of image co-segmentation and also urges the acceptance of standard evaluation. 
	\item Almost all the results on iCoseg, MSRC, and Internet from deep methods surpass the counterparts from traditional methods, except MSRC-14. On FlickerMFC, a non-deep method based on common object models achieved the best result, but actually deep methods haven’t been tested on this dataset. These facts suggest that deep learning is promising to solve the image co-segmentation problem. 
\end{itemize}


\begin{figure*}[htbp]
	
	\centering
	
	\includegraphics{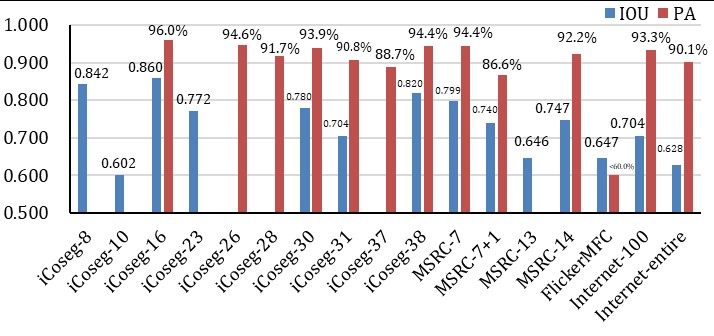}
	
	\caption{Currently best results of PA and IOU over various test sets.}
	
\end{figure*}

\section{Discussions}
\par Image co-segmentation is an important segmentation problem and is potentially useful in many applications. In the last decade, many image co-segmentation methods have been developed and made much progress. However, this is still a very difficult problem. Really good segmentation effects are waiting to be solved. The challenges come from complicated foreground, complicated background, and subtle differences between them. First, an object usually has various visual features, such as various greys, colors, textures, and shapes. Second, the background could be composed of much stuff, e.g., imaging a crowded plaza. Third, the object could be easy confused with the surrounding environment, e.g., imaging an interested person standing in the crowded plaza. Fourth, there are so many classes of objects we could observe; it is almost impossible to model all the possible classes of interested objects. These facts tell that it is very difficult to exactly discriminate the interested objects from the background, especially for handcrafted features and models.
\par Because of its ability of mining sophisticated features from the data, deep learning has shown that it is promising to solve the image co-segmentation problem. However, although it has led to the best results on most of widely used test sets, the improvement is still limited. We think it is helpful to develop more fully end-to-end deep networks by referring to non-deep counterparts. Since many non-deep methods are designed based on graph optimization, while the neural network itself is a graph in structure, it is possible to develop new deep structures by exploring traditional non-deep methods. Along this direction, the number of annotated segmentation data should be considered. Even if PASCAL VOC dataset is obviously bigger than previous datasets, it is still not enough for training satisfactory deep networks. There are three possible ways to solve this problem. First, the weakly supervised learning method can be presented to employ weakly labeled data such as ImageNet. Second, the multi-task learning can be explored to define the learning objective on each pixel in the image, instead of on the whole image, so the training samples can be increased greatly. Third, the reinforcement learning can be employed to get better and better co-segmentation algorithm along with the users’ feedback to the segmentation results.
\par Except deep learning, the correlation between images is another crucial information to the success of image co-segmentation. As described in Section 1, alleviating the ill-pose nature of segmentation by employing the correlation between images is the main advantage of co-segmentation, compared with other segmentation techniques. So we should dig more and better methods to compute the correlation between common objects in different images, at best embedded in deep networks and learned from the data. 
\par Learning objective is an important factor to the success of machine learning based methods. In current image co-segmentation methods based on machine learning, the pixel accuracy (such as popular cross-entropy) is mainly used as the learning objective. The pixel accuracy is not directly related to the quality of segmentation. When the numbers of foreground pixels and background ones are obviously unbalanced, unless the accuracy is very close to 100\%, even seemingly nice accuracy could correspond to a bad segmentation. For solving such foreground/background pixel unbalance problem, we need to seek the learning objective that can directly and more accurately reflect the quality of segmentation. Dice loss\cite{milletari2016v} is a good choice. Actually it has been paid attentions in medical image segmentation in the most recent years\cite{milletari2016v,wong20183d,fidon2017generalised}.

\section{Conclusions}
\par This paper has surveyed the state-of-the-art techniques of fully automatic image co-segmentation algorithms, including traditional methods based on object elements, object regions/contours, and common object models, as well as deep learning based methods. Although much progress has been made, the image co-segmentation is still a very difficult problem. On mainly-used evaluation datasets of iCoseg-38, MSRC-14, Internet-100, and FlickerMFC, the best Intersection Over Union (IOU) values that we achieved are only 0.820, 0.747, 0.704, and 0.647, respectively. The challenges come from complicated foreground, complicated background, and subtle differences between them. Deep learning is a promising direction for solving the problem, under which weakly supervised learning, multi-task learning, reinforcement learning, better correlation computation, and better learning objective are worthy of exploring.

\ifCLASSOPTIONcaptionsoff
  \newpage
\fi



%




\bibliographystyle{IEEEtran}
\bibliography{references}{}

%
\begin{IEEEbiography}[{\includegraphics[width=1in,height=1.25in,clip,keepaspectratio]{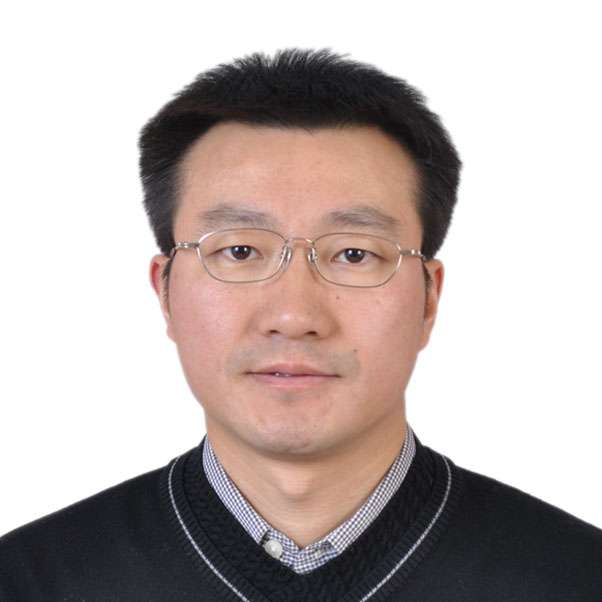}}]{Xiabi Liu}

, Associate professor in School of Computer Science at Beijing Institute of Technology. He received his Ph.D. degree from Beijing Institute of Technology in 2005. His current research interests include medical artificial intelligence, machine learning, computer vision, and pattern recognition (http://www.mlmrlab.com/).
\end{IEEEbiography}

\begin{IEEEbiography}[{\includegraphics[width=1in,height=1.25in,clip,keepaspectratio]{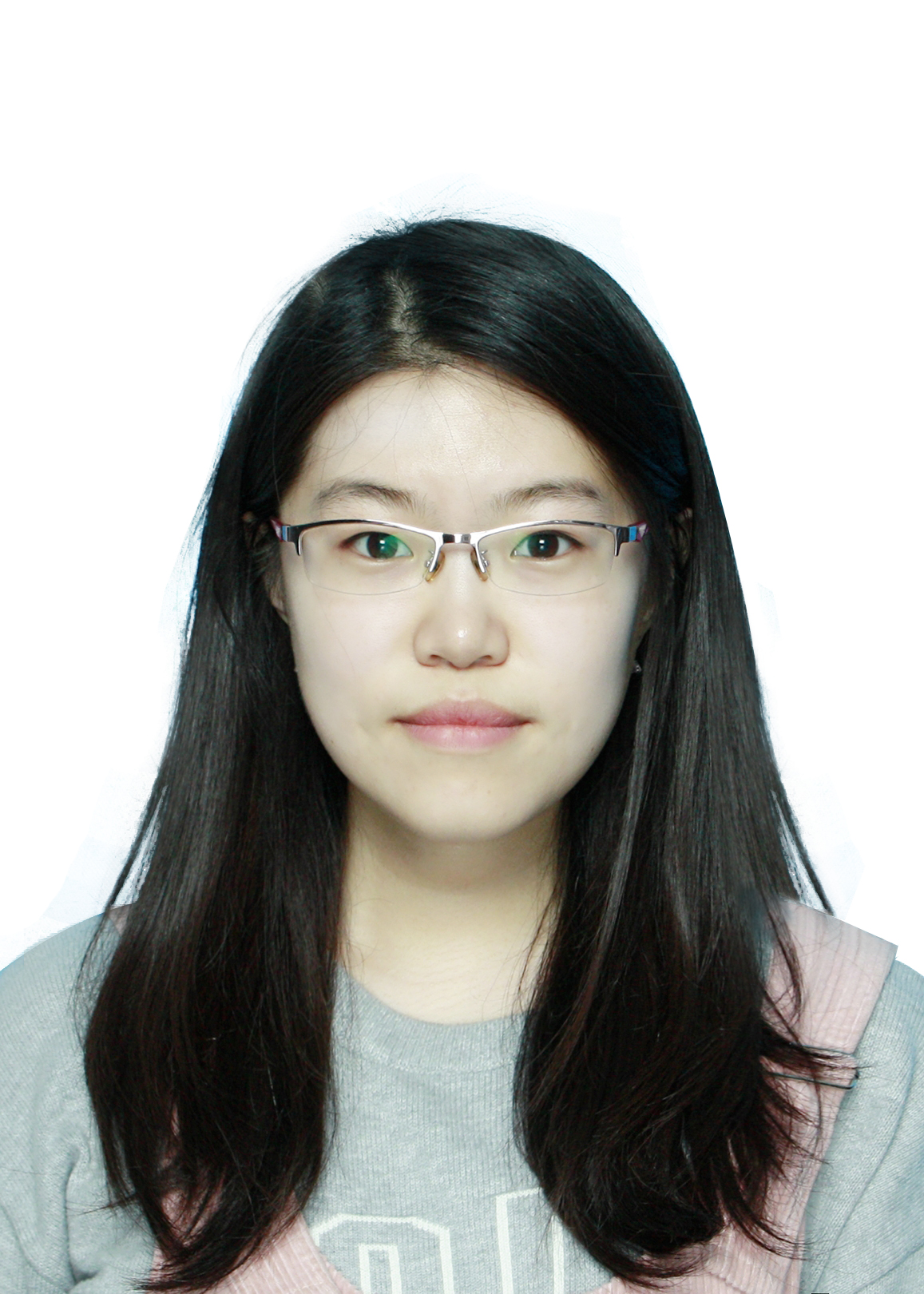}}]{Xin Duan}
was admitted by the school of Computer at Beijing Institute of Technology, Beijing, China. She is currently working toward the master's degree at Beijing Institute of Technology. Her research interests include image co-segmentation and machine learning.
\end{IEEEbiography}





\begin{table*}[!t]
	\renewcommand{\arraystretch}{1.3}
	\caption{COMPARISONS OF PA AND IOU VALUES FOR IMAGE CO-SEGMENTATION, COLLECTED FROM PREVIOUS WORKS}
	\centering
	\begin{tabular}{c||c||c||c||c}
		\hline
		\bfseries	Ref.     & \bfseries Method      & \bfseries Dataset         & \bfseries Criterion & \bfseries Result          \\ \hline\hline
		
		\multirow{2}{*}{\cite{dai2013cosegmentation}}  & \multirow{2}{*}{Common Object Model}     & MSRC-14                          & IOU       & 0.63            \\ \cline{3-5} 
		&                                          & iCoseg-38                        & PA        & 89.50\%         \\ \hline
		{\cite{meng2016cosegmentation}}                  & Object regions/contours                  & iCoseg-38                        & IOU       & 0.71            \\ \hline
		{\cite{joulin2010discriminative}}                   & Object elements                          & MSRC-7                           & PA        & 79.20\%         \\ \hline
		{\cite{kim2011distributed}}                   & Object elements                          & MSRC-13                          & IOU       & 0.646           \\ \hline
		{\cite{lee2015multiple}}                   & Object elements                          & iCoseg-16                        & PA        & 93.10\%         \\ \hline
		\multirow{6}{*}{\cite{yuan2017deep}}  & \multirow{6}{*}{Deep}                    & \multirow{2}{*}{iCoseg-16 }                       & PA        & 96.00\%         \\ \cline{4-5} 
		&                                          &                                  & IOU       & 0.86            \\ \cline{3-5} 
		&                                          & \multirow{2}{*}{iCoseg-38}                        & PA        & 94.40\%         \\ \cline{4-5} 
		&                                          &                                  & IOU       & 0.82            \\ \cline{3-5} 
		&                                          & \multirow{2}{*}{Internet-100}                     & PA        & 91.10\%         \\ \cline{4-5} 
		&                                          &                                  & IOU       & 0.677           \\ \hline
		{\cite{zhu2014multiple}}                  & Common object models                     & FlickerMFC                       & PA        & \textless{}60\% \\ \hline
		\multirow{6}{*}{\cite{wang2017multiple}} & \multirow{6}{*}{Deep}                    & \multirow{2}{*}{iCoseg-38}       & PA        & 93.80\%         \\ \cline{4-5} 
		&                                          &                                  & IOU       & 0.77            \\ \cline{3-5} 
		&                                          & iCoseg-26                        & PA        & 94.60\%         \\ \cline{3-5} 
		&                                          & \multirow{2}{*}{MSRC-14}         & PA        & 90.90\%         \\ \cline{4-5} 
		&                                          &                                  & IOU       & 0.73            \\ \cline{3-5} 
		&                                          & MSRC-7+1                         & IOU       & 0.74            \\ \hline
		{\cite{meng2012object}}                  & Object regions/contours                  & iCoseg-28                        & PA        & 91.70\%         \\ \hline
		\multirow{10}{*}{\cite{han2018robust}} & \multirow{10}{*}{Deep}                   & \multirow{2}{*}{iCoseg-38}       & PA        & 94.40\%         \\ \cline{4-5} 
		&                                          &                                  & IOU       & 0.782           \\ \cline{3-5} 
		&                                          & \multirow{2}{*}{iCoseg-30}       & PA        & 93.90\%         \\ \cline{4-5} 
		&                                          &                                  & IOU       & 0.78            \\ \cline{3-5} 
		&                                          & \multirow{2}{*}{iCoseg-16}       & PA        & 95.80\%         \\ \cline{4-5} 
		&                                          &                                  & IOU       & 0.846           \\ \cline{3-5} 
		&                                          & \multirow{2}{*}{Internet-entire} & PA        & 90.10\%         \\ \cline{4-5} 
		&                                          &                                  & IOU       & 0.628           \\ \cline{3-5} 
		&                                          & \multirow{2}{*}{Internet-100}    & PA        & 90.10\%         \\ \cline{4-5} 
		&                                          &                                  & IOU       & 0.62            \\ \hline
		
		\multirow{2}{*}{\cite{li2016object}} & \multirow{2}{*}{Common object models}    & iCoseg-10                        & IOU       & 0.526           \\ \cline{3-5} 
		&                                          & MSRC-14                          & IOU       & 0.58            \\ \hline
		\multirow{2}{*}{\cite{liu2014object}}  & \multirow{2}{*}{Common object models}    & iCoseg-23                        & IOU       & 0.772           \\ \cline{3-5} 
		&                                          & FlickerMFC                       & IOU       & 0.647           \\ \hline
		\multirow{2}{*}{\cite{rubio2012unsupervised}} & \multirow{2}{*}{Common object models}    & iCoseg-16                        & PA        & 83.90\%         \\ \cline{3-5} 
		&                                          & MSRC-7+1                         & PA        & 70.80\%         \\ \hline
		\multirow{8}{*}{\cite{rubinstein2013unsupervised}} & \multirow{8}{*}{Object regions/contours} & \multirow{2}{*}{iCoseg-30}       & PA        & 89.60\%         \\ \cline{4-5} 
		&                                          &                                  & IOU       & 0.676           \\ \cline{3-5} 
		&                                          & \multirow{2}{*}{MSRC-14}         & PA        & 92.20\%         \\ \cline{4-5} 
		&                                          &                                  & IOU       & 0.747           \\ \cline{3-5} 
		&                                          & \multirow{2}{*}{Internet-entire} & PA        & 84.50\%         \\ \cline{4-5} 
		&                                          &                                  & IOU       & 0.576           \\ \cline{3-5} 
		&                                          & \multirow{2}{*}{Internet-100}    & PA        & 85.40\%         \\ \cline{4-5} 
		&                                          &                                  & IOU       & 0.573           \\ \hline
		\multirow{3}{*}{\cite{liu2017image}} & \multirow{3}{*}{Object elements}         & iCoseg-30                        & IOU       & \textless{}0.7  \\ \cline{3-5} 
		&                                          & MSRC-14                          & IOU       & \textless{}0.7  \\ \cline{3-5} 
		&                                          & FlickerMFC                       & IOU       & \textless{}0.6  \\ \hline
		\cite{chang2011co}                   & Object regions/contours                  & MSRC-7+1                         & PA        & 86.60\%         \\ \hline
		\multirow{2}{*}{\cite{zhang2014image}} & \multirow{2}{*}{Common object models}    & iCoseg-37                        & PA        & 88.70\%         \\ \cline{3-5} 
		&                                          & MSRC-14                          & PA        & 80.60\%         \\ \hline
		\multirow{2}{*}{\cite{liu2014object}}  & \multirow{2}{*}{Common object models}    & iCoseg-16                        & PA        & 85.40\%         \\ \cline{3-5} 
		&                                          & MSRC-7                           & PA        & 90.20\%         \\ \hline
		
	\end{tabular}
\end{table*}	
~\\

\begin{table*}[!t]
	\renewcommand{\arraystretch}{1.3}
	\centering
	\begin{tabular}{c||c||c||c||c}
		
		\hline
		\bfseries	Ref.     & \bfseries Method      & \bfseries Dataset         & \bfseries Criterion & \bfseries Result          \\ \hline\hline
		\multirow{8}{*}{\cite{faktor2013co}} & \multirow{8}{*}{Object elements}         & \multirow{2}{*}{iCoseg-38}       & PA        & 92.80\%         \\ \cline{4-5} 
		&                                          &                                  & IOU       & 0.73            \\ \cline{3-5} 
		&                                          & \multirow{2}{*}{iCoseg-16}       & PA        & 94.40\%         \\ \cline{4-5} 
		&                                          &                                  & IOU       & 0.79            \\ \cline{3-5} 
		&                                          & \multirow{2}{*}{MSRC-14}         & PA        & 89.20\%         \\ \cline{4-5} 
		&                                          &                                  & IOU       & 0.73            \\

		\cline{3-5} 
		&                                          & \multirow{2}{*}{MSRC-7}          & PA        & 92\%            \\ \cline{4-5} 
		&                                          &                                  & IOU       & 0.77            \\ \hline
		\cite{fu2015object}                  & Object regions/contours                  & iCoseg-16                        & PA        & 90.70\%         \\ \hline
		\cite{Jian2013Learning}                  & Common object models                     & MSRC-14                          & IOU       & 0.538           \\ \hline
		\multirow{2}{*}{\cite{liang2017multi}} & \multirow{2}{*}{Object elements}         & iCoseg-38                        & PA        & 87.80\%         \\ \cline{3-5} 
		&                                          & MSRC-14                          & PA        & 78.50\%         \\ \hline
		\cite{kim2012multiple}                  & Common object models                     & FlickerMFC                       & IOU       & \textless{}0.5  \\ \hline
		\cite{wang2013semi}                  & Object regions/contours                  & iCoseg-10                        & IOU       & 0.602           \\ \hline
		\cite{li2018unsupervised}                  & Object elements                          & iCoseg-38                        & IOU       & 0.706           \\ \hline
		\multirow{4}{*}{\cite{tao2017image}} & \multirow{4}{*}{Object elements}         & \multirow{2}{*}{iCoseg-31}       & PA        & 90.80\%         \\ \cline{4-5} 
		&                                          &                                  & IOU       & 0.704           \\ \cline{3-5} 
		&                                          & \multirow{2}{*}{Internet-100}    & PA        & 83.40\%         \\ \cline{4-5} 
		&                                          &                                  & IOU       & 0.548           \\ \hline
		\cite{kim2012hierarchical}                  & Object elements                          & MSRC-13                          & IOU       & 0.475           \\ \hline
		\multirow{2}{*}{\cite{wang2013image}} & \multirow{2}{*}{Common object models}    & iCoseg-16                        & PA        & 90.50\%         \\ \cline{3-5} 
		&                                          & MSRC-7                           & PA        & 87.10\%         \\ \hline
		\multirow{5}{*}{\cite{li2018deep}} & \multirow{5}{*}{Deep}                    & iCoseg-8                         & IOU       & 0.842           \\ \cline{3-5} 
		&                                          & \multirow{2}{*}{MSRC-7}          & PA        & 94.40\%         \\ \cline{4-5} 
		&                                          &                                  & IOU       & 0.799           \\ \cline{3-5} 
		&                                          & \multirow{2}{*}{Internet-100}    & PA        & 93.30\%         \\ \cline{4-5} 
		&                                          &                                  & IOU       & 0.704           \\ \hline
		
	\end{tabular}
\end{table*}

\end{document}